\documentclass[runningheads]{llncs}
\usepackage[table]{xcolor}
\usepackage{graphicx}
\usepackage{amsmath}
\usepackage{array,multirow,booktabs}
\usepackage{amsfonts}

\begin{document}
\title{Rethinking Semi-Supervised Federated Learning: How to co-train fully-labeled and fully-unlabeled client imaging data}
\titlerunning{Rethinking Semi-Supervised Federated Learning}
\authorrunning{Saha et al.}

\author{Pramit Saha\inst{1}, Divyanshu Mishra\inst{1}, J.Alison Noble\inst{1}}
\institute{Department of Engineering Science, University of Oxford}

\maketitle              
\begin{abstract}
The most challenging, yet practical, setting of semi-supervised federated learning (SSFL) is where a few clients have fully labeled data whereas the other clients have fully unlabeled data. This is particularly common in healthcare settings where collaborating partners (typically hospitals) may have images but not annotations. The bottleneck in this setting is the joint training of labeled and unlabeled clients as the objective function for each client varies based on the availability of labels. This paper investigates an alternative way for effective training with labeled and unlabeled clients in a federated setting. We propose a novel learning scheme specifically designed for SSFL which we call Isolated Federated Learning (IsoFed) that circumvents the problem by avoiding simple averaging of supervised and semi-supervised models together. In particular, our training approach consists of two parts - (a) isolated aggregation of labeled and unlabeled client models, and (b) local self-supervised pretraining of isolated global models in all clients. We evaluate our model performance on medical image datasets of four different modalities publicly available within the biomedical image classification benchmark MedMNIST. We further vary the proportion of labeled clients and the degree of heterogeneity to demonstrate the effectiveness of the proposed method under varied experimental settings.

\end{abstract}

\section{Introduction}

Federated Learning (FL) \cite{ji2021emerging,kairouz2021advances,li2020federated,zhang2021survey} is a distributed learning approach that allows the collaborative training of machine learning models using data from decentralized sources while preserving data privacy. However, most current FL methods have limitations, including assuming fully annotated and homogeneous data distribution among local clients. In a practical scenario, like a multi-institutional healthcare collaboration, the participating clients (\textit{i.e.}, medical institutions and hospitals) may not have the incentive or resources to annotate their data \cite{liu2021federated}. To address this, semi-supervised federated learning (SSFL) \cite{diao2021semifl,liu2021federated,zhang2021improving} methods have been proposed to utilize unlabeled data and integrate semi-supervised learning algorithms \cite{van2020survey,berthelot2019mixmatch,sohn2020fixmatch,zhang2021flexmatch,tarvainen2017mean} into federated settings.

Based on the availability of labeled data, the existing SSFL studies can be classified into two main scenarios: (a) labels-at-client, with each client having some labeled and some unlabeled data \cite{jeong2020federated,lin2021semifed}, (b) labels-at-server, with each client possessing only unlabeled data and the server possessing some labeled data \cite{he2021ssfl,jeong2020federated,zhang2021improving,diao2021semifl}. We argue that a more realistic SSFL scenario which is highly challenging but rarely explored in the literature is where some clients have labeled data, and others have completely unlabeled data \cite{liang2022rscfed,yang2021federated,liu2021federated}. 

The classic federated averaging scheme aggregates weights of all labeled and unlabeled client models trained in parallel. The labeled clients typically use cross-entropy-based loss functions while the unlabeled clients primarily use consistency regularization loss \cite{sohn2020fixmatch} or pseudo-labeling-based \cite{arazo2020pseudo,yafen2022survey} semi-supervised learning schemes. This results in high gradient diversity \cite{zhang2021improving} between the supervised and unsupervised models particularly in heterogeneous client settings, as these are targeted to optimize separate objective functions. As a result, the aggregated global model is weak and unable to capture a strong representation of either group of clients. This, in turn, leads to the generation of noisy targets for unlabeled clients and hence the global model fails to converge. The situation is further aggravated under non-IID data distribution conditions where the labeled client class distribution varies greatly from that of unlabeled clients. This naturally poses the following important question: \textit{``How can we effectively co-train supervised and unsupervised models under FL setting that aim to optimize separate objective functions at their respective heterogeneous labeled data or unlabeled data clients?"}

To address this question, we present a novel SSFL algorithm which we call IsoFed that effectively improves client training by isolating the model aggregation of labeled and unlabeled client groups while still leveraging one group of models to improve another. In summary, the primary contributions of this paper are:
\begin{enumerate}

    \item We propose IsoFed, a novel SSFL framework, that realizes isolated aggregation of labeled and unlabeled client models in the server followed by federated self-supervised pretraining of the global model in each individual site. 
    
    \item This is the first work to reformulate model aggregation for fully labeled and fully unlabeled clients under SSFL settings. To the best of our knowledge, we are the first to isolate the aggregation of labeled and unlabeled client models while switching between the two client groups.
    
    \item This work bridges the gap between Federated Learning and Transfer Learning (TL) \cite{weiss2016survey} by combining the best of both worlds for learning across sites. First, we conduct federated model aggregation among the labeled or unlabeled client groups. Next, we leverage Transfer Learning to allow knowledge transfer between the two groups. Therefore, we avoid the issue of averaging the supervised and unsupervised models with high gradient diversity in the context of SSFL while also being unaffected by catastrophic forgetting encountered in multi-domain transfer learning.
    
    \item We, for the first time, extensively evaluate SSFL methods on multiple medical image benchmarks with a varying proportion of clients and degree of heterogeneity. Our results show that the proposed isolated aggregation followed by federated pretraining outperforms the state-of-the-art method, \textit{viz.}, RSCFed \cite{liang2022rscfed} by $\textbf{6.91}\%$ in terms of accuracy and achieves near-supervised learning performance.
\end{enumerate}
\section{Methods}
\begin{figure}[t]

    \centering
\includegraphics[width=1\columnwidth]{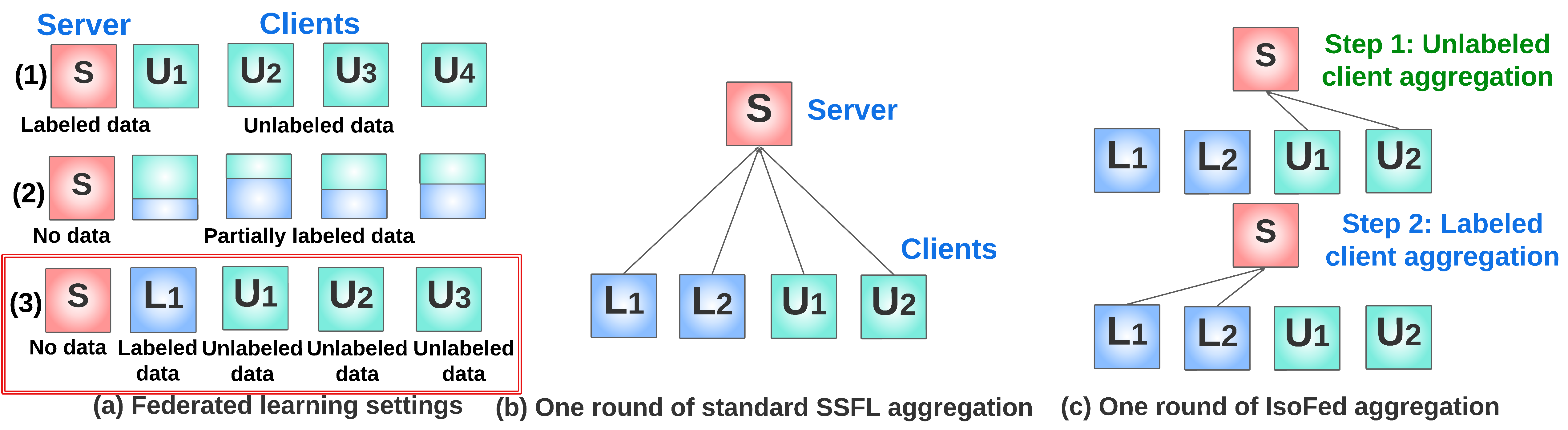}
    \caption{Problem settings and aggregation schemes for semi-supervised federated learning. (a) Three plausible semi-supervised federated learning settings. We address the unique condition (3) with fully labeled and fully unlabeled clients. (b) One round of a standard FL aggregation scheme. (c) One round of our proposed two-step isolated aggregation scheme for labeled clients and unlabeled clients.}
\label{fig:0}
\end{figure}
\subsection{Problem Description}
Assume a federated learning setting with $m$ fully labeled clients denoted as $\{C_1, C_2,...,C_m\}$ each possessing a labeled dataset $D^l = \{(X_i^l, y_i^l)\}_{i=1}^{N^l}$ and $n$ fully unlabeled clients defined as $\{C_{m+1}, C_{m+2},...,C_{m+n}\}$ each possessing an unlabeled dataset $D^u  = \{(X_i^u)\}_{i=1}^{N^u}$. Our objective is to learn a global model $\theta_{glob}$ via decentralized training.
\subsection{Local Training}
We adopt mean-teacher-based semi-supervised learning \cite{tarvainen2017mean,li2020federated,liang2022rscfed} to train each unlabeled client. At the beginning of each round, the global model $W_{glob}$ is used to initialize the teacher model $W_{t}$. At the end of each communicating round, the student model $W_{s}$ is returned to the server as the local model. Each batch of images undergoes two types of augmentations. The teacher model receives weakly augmented data whereas the student model receives strongly augmented data in each local iteration. In order to decrease entropy of model output, the temperature of predictions is further increased via sharpening operation \cite{chen2020simple,berthelot2019mixmatch,goodfellow2016deep,liang2022rscfed} as $\hat{p}_{t,i} = Sharpen ~(p_t, \tau)_i = p_{t,i}^{\frac{1}{\tau}}$\slash${\sum_j p_{t,j}^{\frac{1}{\tau}}}$ where $p_{t,i}$ and $\hat{p}_{t,i}$ denote each element in $p_t$ before and after sharpening, respectively. $\tau$ denotes the temperature parameter.  The student model is trained on the local data ($D^u$) via consistency regularization with the teacher model output. The consistency regularization loss is defined as $\mathcal{L}_{MSE}=\|\hat{p}_t - p_s \|_2^2$ where $\hat{p_t}$ and ${p_s}$ are teacher and student predictions, respectively. $\|.\|_2^2$ denotes $L2$-norm. The student model weights are optimized via backpropagation whereas the teacher model weights are updated by exponential moving averaging (EMA) after each local iteration, as in Eqn. 1:

\begin{equation}
W_{t+1}=\alpha W_s + (1-\alpha) W_t
\end{equation}
where $\alpha$ denotes momentum parameter. We optimize cross-entropy loss for local training on labeled clients defined as $ \mathcal {L}_{CE} = - y_i\log ~ {p}_i$, where $y_i$ denotes labels.


\subsection{Isolated Federated Aggregation}
In this section, we explain the proposed isolated aggregation of labeled and unlabeled client models. Each communication round is composed of two consecutive substeps. First, the server initializes the global model $W_{glob}^t$ and sends it to \textbf{unlabeled} clients ($U_i$). The global model is used to initialize the teacher model $W_t$ in each client. At this stage, only the unlabeled clients perform local training on the global model by minimizing the consistency regularization loss. The updated semi-supervised models obtained after running the local epochs are then uploaded to the server. We adopt a dynamically weighted Federated Averaging scheme \cite{liang2022rscfed} to aggregate the model parameters of all unlabeled clients $W_u$ at the server. For this, we first obtain the averaged model by performing Fed-Avg as in Eqn. 2.   

\begin{equation}
    W_{avg} = \frac{ \sum_{k=1}^{k=K} n_k W_k}{\sum_{k=1}^{k=K} n_k}
\end{equation}
 where $K$ is the total number of clients. $n_k$ is the number of samples in each client. The client models are then dynamically scaled using coefficients $c_k$ designed as functions of the individual distances from the averaged model as denoted in Eqn. 3. The global model ($W_{glob}$) is updated by re-aggregating the client weights scaled by new coefficients $c_k$. In Eqn. 3, $\lambda_{c}$ is a hyperparameter.

\begin{equation}
    c_{k}= \frac{ n_k \exp (-{\lambda_{c}\frac{  { \| W_k - W_{avg}\|}_2}{ n_k}}) }{\sum_{k=1}^{k=K} n_k}, W_{glob} =  \frac{\sum_{k=1}^{k=K} c_k W_k}{\sum_{k=1}^{k=K} c_k}
\end{equation}

The updated global model parameters are then communicated to each \textbf{labeled} client which initializes its models using these weights and trains the local model via minimization of the standard cross-entropy loss. After a pre-defined number of local epochs, each labeled client uploads its local model to the server. The server then aggregates all the supervised models employing the aforementioned weighting scheme and the resultant global model $W_{glob}^{t+1}$ is then sent to each unlabeled client at the beginning of the next round.

\begin{figure}[t]

    \centering
\includegraphics[width=1.05\columnwidth]{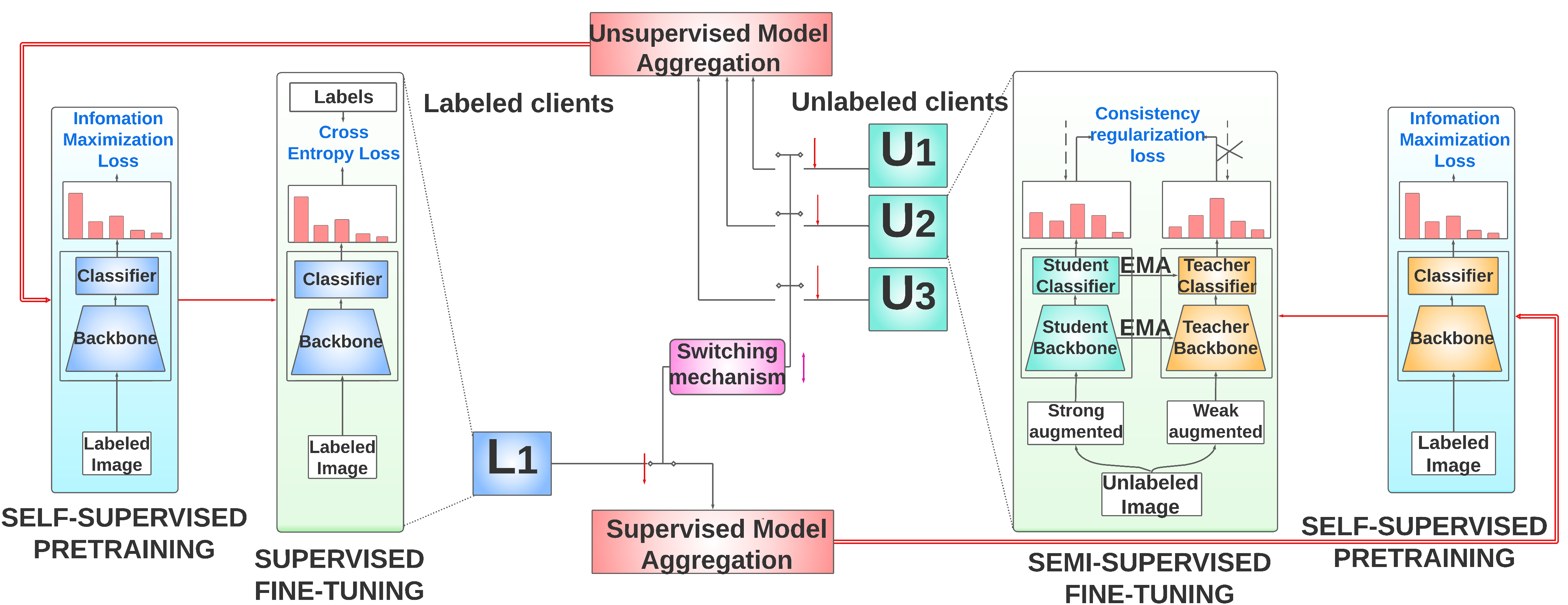}
    \caption{Overview of our proposed methodology (IsoFed) with 1 labeled and 3 unlabeled clients. The unlabeled clients are trained using a mean-teacher-based SSL model. A switching mechanism swaps between labeled and unlabeled clients for isolated model aggregation in each round. After isolated model aggregation, an information maximization loss is used for client-adaptive pretraining to enhance the certainty
and diversity of predictions of the global model for each client before actual local training.}
\label{fig:1}
\end{figure}
\subsection{Client-adaptive Pretraining}
Motivated by the recent success of continued pretraining in Natural Language Processing \cite{gururangan2020don,howard2018universal,liu2019roberta}, we present a client-adaptive pretraining strategy as the second part of our proposed method. If we view the isolated FL from a transfer learning  perspective, the global model received in one group of clients from the server can be regarded as an averaged model pretrained on the other group of clients. To improve client-specific model performance, we conduct a second phase of in-client federated pretraining on the global model before initializing it as a teacher model. 

For self-supervised pretraining, we jointly learn the client-invariant features and client-specific classifier by optimizing an information-theoretic metric called information maximization (IM) loss denoted as $\mathcal{L}_{inf}$ in Eqn. 4. It acts as an estimate of the expected misclassification error of the global model for each client. Optimizing the IM loss makes the global model output predictions that are individually certain but collectively diverse. With the help of a diversity preserving regularizer (first component in Eqn. 4), IM avoids the trivial solution of entropy minimization where all unlabeled data collapses to the same one-hot encoding.  
 The joint optimization is done by reducing the entropy of the output probability distribution of global model ($p_i$) in conjunction with maximizing the mutual information between the data distribution and the estimated output distribution yielded by the global model. 
\begin{equation}
\mathcal{L}_{inf}= \mathbb{E}_{x\in {D}} \left[ \left( \frac{1}{N}\sum_{i=1}^{N} {p_i} \right) \log \left({ \frac{1}{N}\sum_{i=1}^{N} p_i}\right)- \frac{1}{N}\sum_{i=1}^{N} {p_i} \log {p_i} \right]
\end{equation}
where $N$ is the number of classes. $x$ denotes any instance belonging to a dataset $D$. The entropy minimization leads to the least number of confused predictions whereas the regularizer avoids the degenerate solution where every data sample is assigned to the same class \cite{shi2012information,liang2020we}. The pretrained model is then initialized as the teacher model to train the local student model in each round.

\section{Experiments and Results}

\subsection{Datasets and FL settings} To evaluate the performance and generalisability of the proposed method, we conduct experiments on four publicly available medical image benchmark datasets with different modalities \cite{yang2023medmnist}, \textit{viz.}, BloodMNIST (microscopic peripheral blood cell images), PathMNIST (colon pathology), PneumoniaMNIST (chest X-ray), and OrganAMNIST (abdominal CT - axial view). Each image resolution is $28 \times 28$ pixels and is normalized before feeding it to the network. BloodMNIST contains a total of 17,092 images and is organized into 8 classes. PathMNIST has 107,180 images and has 9 types of tissues. PneumoniaMNIST is a collection of 5,856 images and the task is binary classification (diseased vs normal). OrganAMNIST is comprised of 58,850 images and the task is multi-class classification of 11 body organs. We split each training dataset between 4 clients to mimic a practical collaborative setting in healthcare. To testify the versatility of the models, we study two challenging non-IID data partition strategies with 0.5 and 0.8-Dirichlet ($\gamma$). As a result, the number of samples per class and per client widely vary from each other. Additionally, we show the impact of varying the proportion of labeled clients ($75\%, 50\%, 25\%$) on model performance. See \textbf{Suppl. Sec 1} for more details.

\noindent
\subsection{Implementation and training details}
For all datasets, we employ a simple CNN comprising of two 5×5
convolution layers, a 2×2 max-pooling layer, and two fully-connected layers as the feature extraction backbone followed by a two-layer MLP and a fully-connected layer as the classification network. Our model is implemented with PyTorch. We follow the settings prescribed for a training RSCFed to enable a fair comparison. 
See \textbf{Suppl. Sec 2} for more training details.
\begin{table*}[t]

  \centering
  \caption{Comparison with baselines on BloodMNIST and PathMNIST. wFedAvg refers to dynamically weighted Federated averaging. UB implies Upper Bound. MT refers to Mean teacher-based SSL. Acc. and Prec. denote Accuracy and Precision. L and U denote the number of labeled and unlabeled clients respectively.} 
 \scalebox{0.9}
 {
\begin{tabular}{|c|c|c|c|c|c|c|c|c|c|c|c|}
\hline
\multirow{3}{*}{\textbf{Labeling}} & \multirow{3}{*}{\textbf{Method}} & \multicolumn{2}{c|}{\textbf{Client}} & \multicolumn{4}{c|}{\textbf{Metrics} ($\%$)} & \multicolumn{4}{c|}{\textbf{Metrics} ($\%$)} \\

\cline{3-12}
 & &\multirow{2}{*}{L} &\multirow{2}{*}{U} & {\textbf{Acc}.} &{\textbf{AUC}} &{\textbf{Prec}.}&{\textbf{Recall}} & {\textbf{Acc}.} &{\textbf{AUC}} &{\textbf{Prec}.}&{\textbf{Recall}}\\
\cline{5-12}
 &&&&\multicolumn{4}{|c|}{$\gamma = 0.8$ (less non-IID)} &\multicolumn{4}{|c|}{$\gamma = 0.5$ (more non-IID)} \\
 \hline
   \multicolumn{12}{|c|}{\textbf{Dataset 1 : BloodMNIST, Task : 8-class classification}} \\
  \hline
 {Fully supervised} & wFed-Avg (UB)  & 4 & 0 & 79.57&96.61&77.65&75.70&79.45& 96.80 & 78.28 & 73.31 \\
\hline 
 \multirow{3}{*}{} & MT+wFed-Avg & 3 & 1 & 77.32&96.70&74.16&73.79& 70.89 & 95.11& 73.46 & 65.06\\
 & RSCFed  & 3 & 1  & 76.94 & 95.54 & 75.11 & 71.18 & 75.18 & 94.99&76.55&68.96\\
 \cline{2-12}  

 & \textbf{IsoFed}  & 3 & 1 & \textbf{79.43} & \textbf{97.32} & \textbf{76.70} & \textbf{76.67} & \textbf{76.10} & \textbf{95.88} & \textbf{77.13} & \textbf{72.29}\\
\cline{2-12} 

 \multirow{3}{*}{Semi supervised} & MT+wFed-Avg & 2 & 2  & 75.88&96.56&72.85& 71.94& 58.29 & 88.35 & 57.85 & 60.46\\
 & RSCFed  & 2 & 2 & 75.97&95.30&73.58&72.77& 61.18 & \textbf{91.50} & 54.85 & 60.79\\
 \cline{2-12}  

 & \textbf{IsoFed}  & 2 & 2 &\textbf{80.47}&\textbf{97.25}&\textbf{77.11}&\textbf{78.11}&\textbf{64.05}&90.01&\textbf{60.26}&\textbf{64.03}\\
\cline{2-12} 

 \multirow{4}{*}{} & MT+wFed-Avg & 1 & 3 & 75.24 & 95.13 & 72.43 & 70.37 &  52.56 & 89.39 & 57.89 & 55.81\\
 & RSCFed  & 1 & 3 & 71.88 & 93.96 &70.47  & 67.75& 19.35&64.31&07.05& 23.62 \\
 \cline{2-12}  

 & \textbf{IsoFed}  & 1 & 3 & \textbf{79.23} & \textbf{96.43} & \textbf{76.68} & \textbf{77.00} & \textbf{63.70} &\textbf{90.58}&\textbf{70.22}&\textbf{63.81}\\
\hline
  \multicolumn{12}{|c|}{\textbf{Dataset 2 : PathMNIST, Task : 9-class classification}} \\
  \hline
 {Fully supervised} & wFed-Avg (UB)  & 4 & 0 &70.45&94.92&72.13&69.84&68.97 & 94.93 & 68.05 & 67.58   \\
\hline 
 \multirow{3}{*}{} & MT+wFed-Avg & 3 & 1 & 60.97 & 93.60&68.14 &62.00 & 57.92 & 92.93 & \textbf{67.20} & 59.98\\
 & RSCFed  & 3 & 1  &61.55&93.71&61.00&58.95& 58.33 & 93.59 & 60.68 & 58.73\\
 \cline{2-12}  

 &\textbf{IsoFed}  & 3 & 1 & \textbf{63.10} & \textbf{94.73} & \textbf{69.25} & \textbf{64.62} & \textbf{60.23} & \textbf{93.98} & 52.80 & \textbf{61.66}\\
\cline{2-12} 

 \multirow{3}{*}{Semi supervised} & MT+wFed-Avg & 2 & 2  &67.10&\textbf{95.17}&\textbf{66.41}&\textbf{66.4}0&61.28& 91.26 & 61.50 & 57.56\\
 & RSCFed  & 2 & 2 &64.18&93.17&60.79&58.89& 58.83 & 90.35 & 58.88 & 55.02\\
 \cline{2-12}  

 & \textbf{IsoFed}  & 2 & 2 &\textbf{ 70.32} & 94.74 & 65.96 & 64.86 & \textbf{64.00} & \textbf{93.46} & \textbf{63.88} & \textbf{61.22}\\
\cline{2-12} 

 \multirow{3}{*}{} & MT+wFed-Avg & 1 & 3 & 59.57 & 90.66 & 63.14 & 58.93 &  56.31 & 89.92 & 60.42 & 53.92\\
 & RSCFed  & 1 & 3 & 64.75& \textbf{94.09}& \textbf{66.89} & \textbf{63.66} & 57.42 & 89.43 & 54.96 & 53.53\\
 \cline{2-12}  

 & \textbf{IsoFed}  & 1 & 3 & \textbf{66.48} & 92.24 & {63.71} & 62.06 & \textbf{64.02} & \textbf{93.99} & \textbf{66.12} & \textbf{62.39} \\
\bottomrule
\end{tabular}}
\end{table*}%

\begin{table*}[t]

  \centering
  \caption{Performance comparison of IsoFed with baselines on PneumoniaMNIST and OrganAMNIST (with ablation study). PT refers to the federated pretraining step.} 
 \scalebox{0.9}{
\begin{tabular}{|c|c|c|c|c|c|c|c|c|c|c|c|}
\hline
\multirow{3}{*}{\textbf{Labeling}} &\multirow{3}{*}{\textbf{Method}} & \multicolumn{2}{c|}{\textbf{Client}} & 
    \multicolumn{4}{c|}{\textbf{Metrics} ($\%$)}&\multicolumn{4}{c|}{\textbf{Metrics} ($\%$)}\\
\cline{3-12}
 & &\multirow{2}{*}{L} &\multirow{2}{*}{U} & {\textbf{Acc}.} &{\textbf{AUC}} &{\textbf{Prec.}}&{\textbf{Recall}} & {\textbf{Acc}.} &{\textbf{AUC}} &{\textbf{Prec}.}&{\textbf{Recall}}\\
 
\cline{5-12}
 &&&&\multicolumn{4}{|c|}{$\gamma = 0.8$ (less non-IID)} &\multicolumn{4}{|c|}{$\gamma = 0.5$ (more non-IID)} \\
 \hline
   \multicolumn{12}{|c|}{\textbf{Dataset 3 : PneumoniaMNIST, Task : Binary classification}} \\

 \hline
 {Fully supervised} & wFed-Avg (UB)  & 4 & 0 &87.34&95.32&86.71 &89.02  & 87.02 &95.64&86.45&88.76  \\
\hline \cline{2-12}
 \multirow{3}{*}{} & MT+wFed-Avg & 3 & 1 &86.54&95.20&85.94&88.21& 86.86& 94.85 & 85.92  & 87.86  \\
 & RSCFed  & 3 & 1  & 86.58&95.63&\textbf{89.02}&88.68& 86.70& 94.50 &85.75  &87.65  \\
 \cline{2-12}  

 & \textbf{IsoFed}  & 3 & 1 & \textbf{87.10} & \textbf{95.04}& {86.45}& \textbf{89.00} & \textbf{89.26} & \textbf{95.80}& \textbf{88.26} & \textbf{89.44} \\

\cline{2-12} \cline{2-12}

 \multirow{3}{*}{Semi supervised} & MT+wFed-Avg & 2 & 2  &83.65&89.74&82.45&82.99& 82.21 & 96.17  & 83.20  & 85.26\\
 & RSCFed  & 2 & 2 &78.37&87.36&77.31&78.76& \textbf{84.46} & \textbf{95.58}  &\textbf{84.58}  &\textbf{86.88}\\
 \cline{2-12}  

 & \textbf{IsoFed}  & 2 & 2 & \textbf{84.70} & \textbf{90.75} & \textbf{83.56} & \textbf{84.64} &82.68&95.15&83.34&85.41\\

\cline{2-12} \cline{2-12}

 \multirow{3}{*}{} & MT+wFed-Avg & 1 & 3 &81.41&89.84&82.05&77.69& \textbf{79.97}&\textbf{94.45}  &\textbf{81.28}  & \textbf{83.12}\\
 & RSCFed  & 1 & 3 &78.85&86.66&77.56&76.84& 62.50 & 50.00 & 31.25 & 50.00 \\
 \cline{2-12}  

 & \textbf{IsoFed}  & 1 & 3 & \textbf{85.00} & \textbf{91.68} & \textbf{83.98} & \textbf{83.95} & 77.12 & 93.65 & 80.47 & 81.40  \\


\hline
  \multicolumn{12}{|c|}{\textbf{Dataset 4 : OrganAMNIST, Task: 11-class classification}} \\

 \hline
 {Fully supervised} & wFed-Avg (UB)  & 4 & 0 &69.72&94.41&67.44&69.60& 69.50 & 94.63  &68.12  & 69.60   \\
\hline \cline{2-12}
 \multirow{3}{*}{} & MT+wFed-Avg & 3 & 1 &68.36&93.72&68.02&69.38& 66.49  & 93.69 & 67.51 & 68.25 \\
 & RSCFed  & 3 & 1  &68.14&94.26&67.44&69.53& 67.08&93.82  & 68.82 & 68.36 \\
 \cline{2-12}  
& {IsoFed w/o PT}  & 3 & 1 & 68.98 & 94.32 & \textbf{68.83} & 69.88 & 67.45 & 93.98 & 67.85 & 69.35\\
 & \textbf{IsoFed}  & 3 & 1& \textbf{69.47} & \textbf{95.05} & {68.04} & \textbf{70.85} & \textbf{68.65} & \textbf{94.88} & \textbf{68.64} & \textbf{69.77}\\

\cline{2-12} \cline{2-12}

 \multirow{3}{*}{Semi supervised} & MT+wFed-Avg & 2 & 2  &66.28&92.77&66.12&67.63& 61.71 & \textbf{92.55} & \textbf{65.79}  & 62.66  \\
 & RSCFed  & 2 & 2 &66.68&92.42&66.90&66.56&62.51 &91.89  &64.09  &63.35  \\
 \cline{2-12}  
& {IsoFed w/o PT}  & 2 & 2 & 68.67 & 93.25 & 67.65 & 68.50 & \textbf{64.37} & 92.11 & 65.70 & 65.17\\
 & \textbf{IsoFed}  & 2 & 2& \textbf{68.95} & \textbf{93.95} & \textbf{68.32} & \textbf{69.83} & {64.08} & {92.45} & 64.56 & \textbf{65.47}\\

\cline{2-12} \cline{2-12}

 \multirow{3}{*}{} & MT+wFed-Avg & 1 & 3 &57.75&90.95&61.50&55.68& 50.84 & 87.65  & 60.07  & 48.51  \\
 & RSCFed  & 1 & 3 &58.50&90.86&63.48&55.76& 54.90 & 89.58  & 50.53  & 53.41  \\
 \cline{2-12}  
& {IsoFed w/o PT}  & 1 & 3 & 62.03 & 91.36 & 64.50 & 61.44 &56.40&89.79&61.61&55.72\\
 & \textbf{IsoFed}  & 1 & 3& \textbf{62.77} & \textbf{91.48} & \textbf{64.52} & \textbf{61.79} & \textbf{61.90} & \textbf{91.55} & \textbf{62.39} & \textbf{60.21} \\

\bottomrule

\end{tabular}}

\end{table*}%

\subsection{Results and discussion}
We use the standard metrics - accuracy, area under a ROC curve (AUC), Precision, and Recall to evaluate performance. We observe that the dynamically weighted version of Fed-Avg (discussed in Sec 2.3) outperforms standard Fed-Avg and hence use it as a baseline in this paper instead of vanilla Fed-Avg. In order to fairly evaluate IsoFed, we compare with the following state-of-the-art SSFL benchmarks: (a) MT+wFed-Avg: a combination of Mean Teacher and dynamically weighted Fed-Avg, (b) RSCFed: Random sampling consensus-based FL \cite{liang2022rscfed}. Since RSCFed has already been shown to significantly outperform FedIRM \cite{liu2021federated} and Fed-Consist \cite{yang2021federated} on multiple datasets, we exclude those methods from our comparative study due to space constraints. We consider fully-supervised FL as an upper bound and report the results for both the non-IID settings on each dataset.
 Tables 1-2 show that overall, IsoFed outperforms RSCFed by $6.91\%$, $4.15\%$, $7.28\%$, and $6.71\%$ in terms of average accuracy, AUC, Precision, and Recall respectively.  
 
 Table 1 shows our method and our baselines on 8-class classification with BloodMNIST. L and U denote the number of labeled and unlabeled clients respectively. The average accuracy for fully-supervised FL is $79.51\%$. Among the baselines, MT+wFed-Avg has a higher overall accuracy score of $68.36\%$ while RSCFed has an accuracy score of $63.41\%$. Particularly, we find RSCFed collapses under the most extreme case of $\gamma = 0.5 $ and U=3. IsoFed improves the accuracy score to $73.83\%$ and is stable for all evaluated conditions. Table 1 further reports performance on 9-class classification with PathMNIST. The fully-supervised FL achieves an overall accuracy of $69.71\%$. The baselines have very similar accuracy scores of $60.53\%$ and $60.84\%$ respectively. IsoFed improves it to $64.69\%$. 
 
 Table 2 shows binary classification results on PneumoniaMNIST. The fully-supervised FL has an overall accuracy of $87.18\%$. MT+wFed-Avg and RSCFed achieve average accuracy scores of $83.44\%$ and $79.57\%$. IsoFed has the best accuracy of $85.45\%$. 
 Furthermore, the results of 11-class anatomy classification task on OrganAMNIST are also reported in Table 2. The upper bound accuracy is $69.61\%$ and the baseline accuracies are $61.91\%$ and $62.97\%$ respectively. IsoFed achieves an overall accuracy score of $65.97\%$. 
In general, the performance of all methods decreases with $\gamma$ changing from $0.8$ to $0.5$. It is expected as the clients become more label-skewed due to higher non-IID data partition. However, our approach is least affected by this which is reflected in its accuracy decrease by $2.19\%$ as opposed to $4.45\%$ and $2.94\%$ incurred by baselines. As foreseen, performance also deteriorates with decrease in the number of labeled clients. For L:U = 3:1, 2:2, 1:3, the baseline accuracies degrade by $2.16\%$, $5.61\%$, $15.31\%$ and $2\%$, $5.01\%$, $12.91\%$ w.r.t. fully supervised FL setting. However, for IsoFed, the decrease in accuracy is only $0.55\%$, $3.09\%$,and $7.28\%$, respectively. This proves the near-supervised learning performance of the proposed training method.

The superior performance of IsoFed over the baselines and closer performance to the upper bound demonstrates better learning and generalization. This is achieved by the isolated aggregation strategy and federated pretraining on all datasets.

\noindent
\subsection{Ablation study}
Owing to space constraints, we show ablation experiments only on OrganAMNIST, which provides the most challenging classification task, to evaluate the impact of IsoFed components. (More results in \textbf{Suppl. Sec 2}). Table 2 demonstrates that client-adaptive pretraining improves model accuracy by $5.50\%$ for the most extreme condition of $\gamma=0.5$ and L:U=1:3. 

\section{Conclusion}
We have introduced a novel SSFL framework called IsoFed, an isolated federated learning technique, to address joint training of labeled and unlabeled clients in the context of decentralized semi-supervised learning. It opens a new research direction in learning across domains by unifying two dominant approaches - Federated Learning (among labeled or unlabeled clients) and Transfer Learning (between labeled and unlabeled clients). 
Our results challenge the conventional strategy of co-training fully labeled and fully unlabeled clients in SSFL. Experimental results on 4 different medical imaging datasets with varied proportion of labeled clients ($25, 50, 75 \%$) and varied non-IID distribution (0.5 $\&$ 0.8-Dirichlet) show that IsoFed achieves a considerable boost compared to current state-of-the-art SSFL method. 
IsoFed can be easily incorporated into other federated learning-based aggregation schemes as well as used in conjunction with any other semi-supervised learning framework in federated learning setting. 

\section{Acknowledgement}

This work was supported in part by the UK EPSRC (Engineering and Physical Research Council) Programme Grant EP/T028572/1 (VisualAI), a  UK EPSRC Doctoral Training Partnership award, and the InnoHK-funded Hong Kong Centre for Cerebro-cardiovascular Health Engineering (COCHE) Project 2.1 (Cardiovascular risks in early life and fetal echocardiography).

\bibliographystyle{splncs04}
\bibliography{ref}

\end{document}